\documentclass[sigconf]{acmart}

\makeatletter
\let\@authorsaddresses\@empty
\makeatother

\acmSubmissionID{379}

\usepackage{booktabs} %
\usepackage{makecell} 
\usepackage{multirow}
\usepackage{stfloats}
\usepackage{float}

\usepackage[ruled]{algorithm2e} %

\SetAlFnt{\small}
\SetAlCapFnt{\small}
\SetAlCapNameFnt{\small}
\SetAlCapHSkip{0pt}

\acmJournal{TOG}
\acmConference[SA '22 Conference Papers]{SIGGRAPH Asia 2022 Conference Papers}{December 6--9, 2022}{Daegu, Republic of Korea}
\acmBooktitle{SIGGRAPH Asia 2022 Conference Papers (SA '22 Conference Papers), December 6--9, 2022, Daegu, Republic of Korea}

\acmDOI{10.1145/3550469.3555401}
\acmISBN{978-1-4503-9470-3/22/12}

\begin{document}
\title{Reconstructing Hand-Held Objects from Monocular Video}

\author{Di Huang}
\email{dihuanginfo@gmail.com}
\affiliation{%
 \institution{The University of Sydney}
 \country{Australia}
 }

\author{Xiaopeng Ji}
\email{xp.ji@cad.zju.edu.cn}
\affiliation{%
 \institution{Zhejiang University}
 \country{China}
}

\author{Xingyi He}
\email{hexingyi8@gmail.com}
\affiliation{%
 \institution{Zhejiang University}
 \country{China}
}

\author{Jiaming Sun}
\email{suenjiaming@gmail.com}
\affiliation{%
 \institution{Image Derivative Inc.}
 \country{China}
}

\author{Tong He}
\email{tonghe90@gmail.com}
\affiliation{%
 \institution{Shanghai AI Laboratory}
 \country{China}
}

\author{Qing Shuai}
\email{s_q@zju.edu.cn}
\affiliation{%
 \institution{Zhejiang University}
 \country{China}
}

\author{Wanli Ouyang}
\email{wanli.ouyang@sydney.edu.au}
\affiliation{%
 \institution{Shanghai AI Laboratory}
 \institution{The University of Sydney}
 \country{Australia}
}

\author{Xiaowei Zhou}
\email{xwzhou@zju.edu.cn}
\authornote{Corresponding author.}
\affiliation{%
 \institution{State Kay Lab of CAD\&CG}
 \institution{Zhejiang University}
 \country{China}
}

\renewcommand\shortauthors{Huang, D. et al}

\begin{abstract}
This paper presents an approach that reconstructs a hand-held object from a monocular video. In contrast to many recent methods that directly predict object geometry by a trained network, the proposed approach does not require any learned prior about the object and is able to recover more accurate and detailed object geometry. The key idea is that the hand motion naturally provides multiple views of the object and the motion can be reliably estimated by a hand pose tracker. Then, the object geometry can be recovered by solving a multi-view reconstruction problem. 
We devise an implicit neural representation-based method to solve the reconstruction problem and address the issues of imprecise hand pose estimation, relative hand-object motion, and insufficient geometry optimization for small objects.
We also provide a newly collected dataset with 3D ground truth to validate the proposed approach.
The dataset and code will be released at \url{https://dihuangdh.github.io/hhor}.

\end{abstract}

\ccsdesc[500]{Computing methodologies~Reconstruction}

\keywords{Object reconstruction, joint hand-object reconstruction}

\begin{teaserfigure}
\begin{center}
  \includegraphics[width=1.0\linewidth, trim=0 20 0 60]{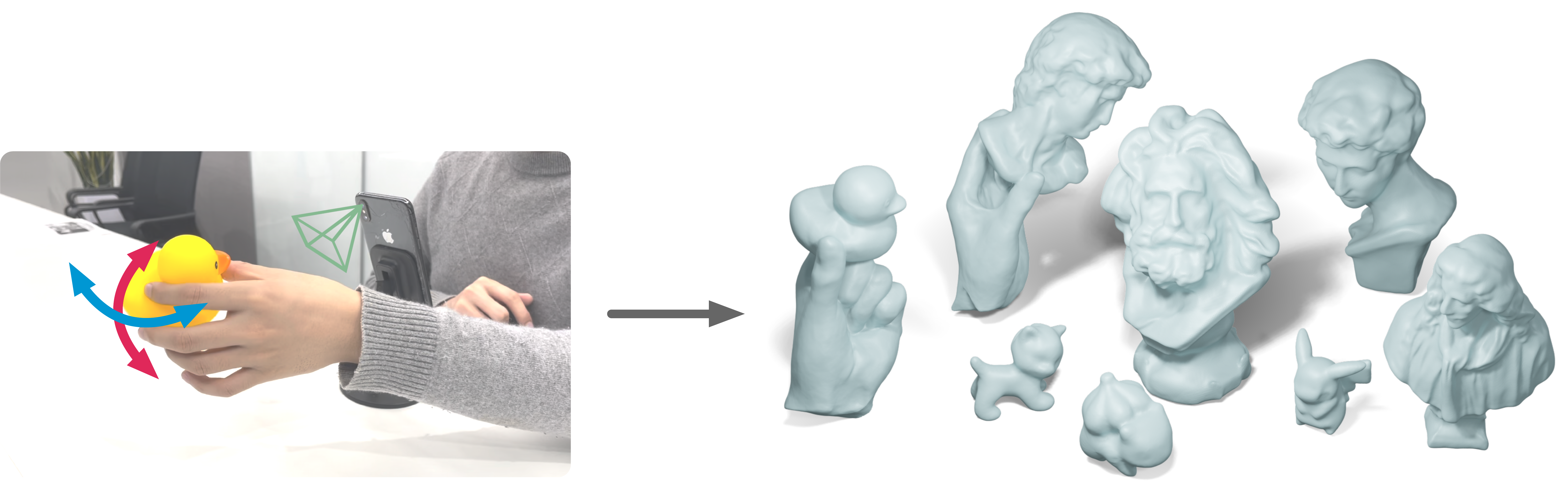}
  \text{\hspace{23mm}\textbf{Capture process}\hspace{72mm}\textbf{Output 3D meshes}\hspace{40mm}}
  \caption{{\textbf{Hand-held object reconstruction.}}
	\textnormal{We propose a novel approach that reconstructs a moving hand-held object from a video captured by a static RGB camera, which is a common real-life scenario. The proposed approach does not require any prior knowledge about the object and thus can be widely applicable. The data capture process (left) and reconstructed 3D meshes (right) are visualized. }
	 }
\label{fig:teaser}
\end{center}
\end{teaserfigure}

\maketitle

\section{Introduction}

Reconstructing a 3D object from 2D images is vital for many applications such as augmented reality, 3D printing, and robotic manipulation. 
Existing approaches \cite{schoenberger2016mvs, izadi2011kinectfusion} mainly focus on the setting of capturing a static object with an RGB camera or depth sensor rotating around the object. 
This work investigates a different setting where the object is held by a moving hand with a fixed grasping gesture in front of a static RGB camera, as shown in Figure \ref{fig:teaser}. This setting is not only common in daily scenarios, e.g., video conferencing, but also potentially provides a more user-friendly object capture procedure without the need to walk around the object. 

The hand-held object capture leads to the following challenges that make existing reconstruction methods inapplicable. 
First, relative motion between the object and the camera needs to be solved for video-based 3D reconstruction. Existing systems generally rely on structure from motion (SfM) algorithms \cite{schoenberger2016sfm}. However, SfM assumes rigid scenes while in our setting, the object is moving independently of the background. Another alternative is to track the 6DoF pose of the object. But traditional object pose estimation methods either need object CAD models \cite{tjaden2016real, tjaden2017real} or rely on rich textures for reliable feature tracking \cite{lowe2004distinctive}, both of which cannot be satisfied in our setting. Recent works \cite{wang2019normalized, chen2020learning} adopt learning-based methods for object pose estimation but need training on the same object category. Second, even if the object motion is given, dense reconstruction of a textureless object from RGB images is still difficult, particularly when the object mask is not provided. 
Moreover, the object is partially occluded by the hand, which makes object pose tracking and reconstruction even harder. 
To solve these challenges, some recent works resort to a learning-based approach that learns object shape prior to making it possible to estimate object shape from a single view with neural networks \cite{karunratanakul2020grasping, hasson19_obman, choy20163d, fan2017point}.
While these works show promising results, the learned single-view reconstruction networks can hardly generalize to new object categories.

In this paper, we propose a novel framework to reconstruct a moving hand-held object from a monocular video without the need to know any prior of the object. The key insight is that the physical constraint between hand and object provides important motion cues for reconstructing the object: while the 3D motion of the object is hard to track, we are able to infer object motion by tracking hand motion based on a learning-based hand pose estimator.
For dense geometry reconstruction, instead of using traditional multi-view stereo algorithms that cannot handle textureless objects, we propose to leverage the recent advances in neural representation-based methods \cite{mildenhall2020nerf,wang2021neus}, which directly optimize 3D scene geometry and appearance represented by implicit functions with differentiable rendering. 
However, directly using such a representation leads to poor reconstruction quality.
We find three main issues that degrade the reconstruction quality: 1) imprecise hand pose, 2) relative motion between the hand and object, and 3) inefficient optimization for the object geometry due to its small size compared to the hand. 
We propose three solutions to address these challenges. 
First, we simultaneously optimize the neural scene representation and the relative pose between the hand and camera. 
Second, we equip the neural scene representation with a learnable deformation field for modeling the relative motion between the hand and object. 
Third, we leverage 2D object masks to adaptively sample more rays on the object during training to make the network more focused on the object instead of the hand. 
Note that by taking an additional background image, the object mask can be obtained using background subtraction and hand segmentation, without the need to know the object category.  
 
To validate the proposed approach, we collect a new hand-held object dataset that consists of 4K videos of 35 objects, 14 of which are
paired with ground-truth meshes obtained by a commercial 3D scanner. 
This dataset covers various types of daily objects.
The experiments show that our approach is able to accurately reconstruct objects and hands and outperforms several baseline methods. 

\section{RELATED WORK}

\begin{figure*}[htb]
\begin{center}
  \setlength{\abovecaptionskip}{0.3cm}
  \includegraphics[width=1.0\linewidth, trim=0 20 0 40]{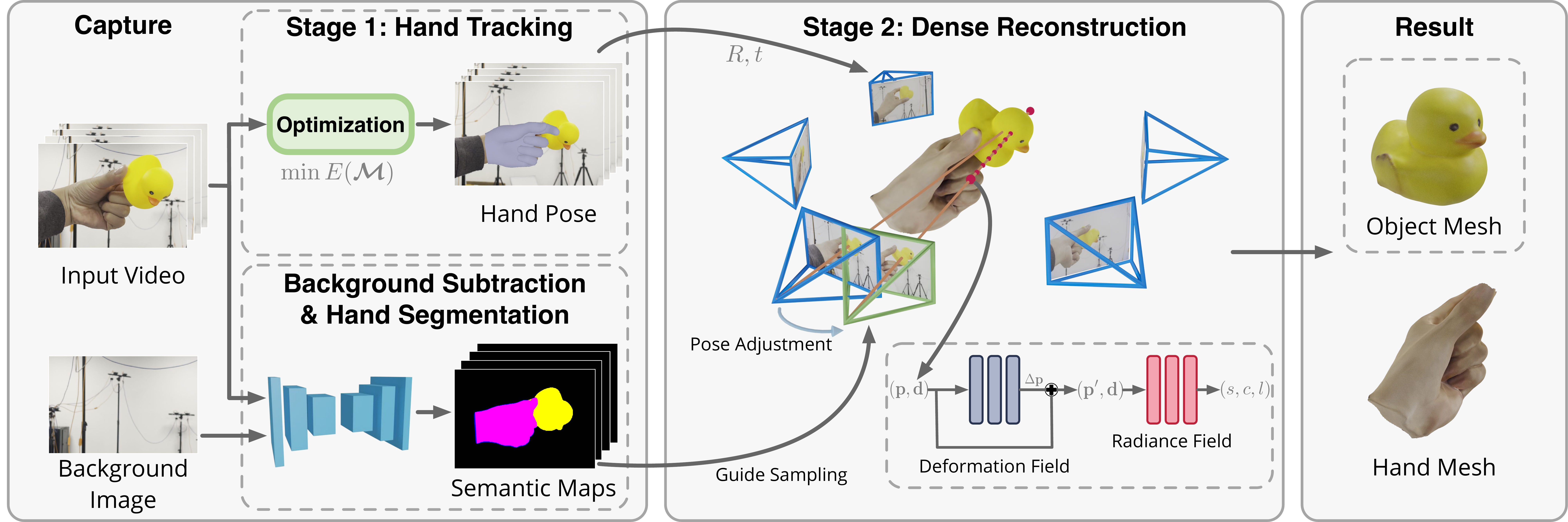}
  \caption{{\textbf{The pipeline of our approach,}}
	\textnormal{which consists of two stages. \textbf{Hand tracking:} by minimizing the reprojection error of hand keypoints detected by a learned detector, the 3D hand pose and the camera motion relative to it are recovered. \textbf{Dense reconstruction}: an implicit neural representation-based method is employed to reconstruct the SDF and color fields of the hand and object. 
	Three additional modules are proposed: the pose adjustment to compensate for imprecise hand pose tracking, the deformation field to model the relative motion between the hand and object, and the semantics-guided sampling to improve object reconstruction quality.	}
	}
\label{fig:pipeline}
\end{center}
\end{figure*}

\paragraph{Multi-view 3D Reconstruction.}
Conventional multi-view 3D reconstruction methods \cite{kutulakos2000theory, furukawa2009accurate, yao2018mvsnet} follow a two-stage pipeline: They first estimate camera parameters by SfM and then use multi-view stereo (MVS) to reconstruct the scene from calibrated images. 
A representative system in this category is COLMAP \cite{schoenberger2016sfm, schoenberger2016mvs}, which first estimates multi-view depth maps and then fuses depth data into 3D models.
There is a recent trend to solve the multi-view 3D reconstruction by recovering the implicit scene representation using volume rendering.
NeRF \cite{mildenhall2020nerf} proposes to represent the scene as a radiance field of density and color. 
NeuS \cite{wang2021neus} improves the NeRF representation by replacing the density radiance field with the Signed Distance Field (SDF), leading to much better reconstruction quality.
These works still require camera poses from SfM, while some more recent works \cite{wang2021nerfmm, lin2021barf} aim to get rid of the SfM phase by simultaneously optimizing the radiance field and camera poses.
However, they are prone to local optimum without proper camera pose initialization \cite{wang2021nerfmm}.
Existing multi-view 3D reconstruction methods mentioned above generally rely on SfM to estimate camera parameters for each frame. 
Unfortunately, SfM cannot work in our setting which aims to reconstruct moving hand-held objects from a monocular RGB video.
First, standard SfM assumes rigid scenes, while in our setting, the object is moving independently of the background. Second, the hand and object are often low-textured, causing standard SfM to fail.

\paragraph{Single-view 3D Reconstruction.}
Another line of works aims at reconstructing 3D objects from single-view input by learning object shape prior. According to the different types of representations, single-view 3D reconstruction methods can be categorized as voxel-based \cite{choy20163d, tatarchenko2017octree, riegler2017octnet, dai2017shape, wang2017shape}, point-based \cite{fan2017point, insafutdinov18pointclouds, mandikal20183d, jiang2018gal}, and mesh-based \cite{ wang2018pixel2mesh, groueix2018}. Voxels are suitable for CNNs to process but suffer from limited resolution. Point clouds and meshes are compact representations, but difficult to be generated by neural networks.
To overcome the limitations of above 3D representations, recent works propose to leverage implicit neural representations \cite{Park_2019_CVPR,chen2019learning, mescheder2019occupancy, NIPS2019_8340,chibane2020neural}. In these methods, the geometry and appearance of a 3D scene are represented as continuous functions of spatial locations, which are approximated by learned neural networks. 
All the above single-view 3D reconstruction methods need to learn strong object shape prior from massive training data. Given the fact that there is a lack of large-scale 3D object datasets with real images and corresponding 3D models, previous methods are often trained on synthetic datasets like ShapeNet \cite{shapenet2015} and suffer from limited generalization capability, particularly when the object category of the test image is unseen before. Moreover, the quality of single-view reconstruction is inherently inferior to multi-view methods.

\paragraph{Hand Tracking}
Hand tracking is an active research topic in computer vision. 
Early works focus on tracking hand motion with a depth sensor. 
\cite{tompson2014real} uses a deep convolutional network to extract features and then apply inverse kinematics for accurate hand pose. 
\cite{taylor2016efficient} presents a hand tracking system for real-time AR applications. The system uses a smooth hand model and non-linear optimization to achieve impressive hand tracking performance. 
\cite{mueller2019real} proposes a novel model which is capable of tracking two interacting hands in real-time.
\cite{hampali2020honnotate} jointly estimates the hand pose and 6D object pose using one or several RGB-D cameras and differentiable rendering.
Recently, hand tracking from RGB input has been paid more attention. \cite{mueller2018ganerated, han2020megatrack} split the hand tracking problem into two stages. They estimate the 3D keypoints locations first and then apply a skeleton fitting process to get the final tracking results.
Other works\cite{ge2019handshapepose, boukhayma20193d, Kulon_2020_CVPR} track the hand by direct estimating pose and shape parameters of a parametric hand model, e.g., the MANO model \cite{MANO:SIGGRAPHASIA:2017}. 

\paragraph{In-hand Capture}
Since hands frequently interact with objects, reconstructing objects in hands is an essential problem for in-hand AR/VR applications. 
Early in-hand scanning uses depth images as input. 
\cite{rusinkiewicz2002real} proposes an in-hand 3D model acquisition system, which allows the user to manipulate the object and see the scanning mesh in real time. The system captures local surface patches, then aligns and integrates them into a complete 3D mesh.
Later work \cite{weise2008accurate} improves the system with efficient registration failure detection and three better registration methods. 
Since the system heavily relies on object textures and often fails for textureless objects, \cite{tzionas20153d} proposes to use the hand contact points as the additional registration energy. 
\cite{zhang2019interactionfusion} proposes a novel pipeline to simultaneously reconstruct hand pose and object shape using two RGBD cameras, which is later simplified to one RGBD camera by \cite{zhang2021single}.
Recent works focus on the simultaneous reconstruction of hand and object from a single RGB image.
\cite{hasson19_obman} jointly reconstructs the hand and objects by using two separate branches: one branch estimates MANO parameters, and the other reconstructs the object shape in the normalized coordinate space. 
\cite{karunratanakul2020grasping, ye2022hand} use implicit representations to recover the object shape.
However, all learning-based methods are limited to a few known object categories included in the training set. 
In contrast, our method works on RGB input and does not rely on any learned object prior to recover high-fidelity object meshes.

\section{DATA CAPTURE and PRE-PROCESSING}

\subsection{Data Capture}
\label{sec:data capture}
Our system only uses a still monocular RGB camera as the capture device (e.g., iPhone or iPad). We first use the camera to take a snapshot of the background scene, which is used to generate the foreground mask later. Then, we capture a short grasping video for hand-object reconstruction.
During the data capture, users only need to grasp and rotate the object in front of the camera. 
The camera is fixed during the capture and the object is firmly grasped by the hand, which means that the relative motion between the hand and the object is small. 
Each video contains around 1800 frames with a 30 fps frame rate and 4K (2160 $\times$ 3840) resolution.

\subsection{Hand-object Segmentation}
Our method requires hand-object segmentation maps to help separate the hand and object. 
We design a two-step framework to generate 2D semantic maps. 
First, we leverage background matting \cite{BGMv2} to extract the foreground masks, which contain both the hand and object. 
Second, we segment the hand from the foreground masks with a hand segmentation network. 
Notably, we collect a new egocentric-view hand segmentation dataset to train the network.
The whole segmentation process does not rely on any object prior and can be widely generalized to various types of objects.

\section{Methods}
This section describes the proposed approach for reconstructing a hand-held object from a monocular RGB video without any prior of the object. 
The proposed approach consists of two main stages: hand tracking (Section~\ref{sec:hand tracking}) and dense reconstruction (Section~\ref{sec:dense reconstruction}). 
An overview of our approach is presented in Figure \ref{fig:pipeline}.

\subsection{Hand Tracking}
\label{sec:hand tracking}

We build our hand tracking method on the widely used parametric hand model MANO~\cite{MANO:SIGGRAPHASIA:2017}.
The hand mesh is defined by two sets of parameters: pose ($\boldsymbol{\theta}$) and shape ($\boldsymbol{\beta}$). The pose parameters affect the joint angles except for the wrist joint, and the shape parameters control the person-specific shape deformations. 
$\boldsymbol{R} \in \mathbf{SO}(3)$ and $\boldsymbol{T} \in \mathbb{R}^{3}$ are the relative rotation and translation between the hand and canonical MANO space, which also indicate the relative motion between the hand and camera. 
$\boldsymbol{R}$, $\boldsymbol{T}$, $\boldsymbol{\theta}$ and $\boldsymbol{\beta}$ share the same parametrization with MANO.
We denote the optimized hand model in this article as $\boldsymbol{M}= \{ \boldsymbol{\theta}, \boldsymbol{\beta}, \boldsymbol{R}, \boldsymbol{T}\}$. The paired joint locations of the MANO model can be denoted as $J(\boldsymbol{M})$.

Given a set of RGB images $\mathcal{I}$ captured by the monocular camera, hand tracking aims to fit the hand model parameters $\boldsymbol{\mathcal{M}}=\{\boldsymbol{M}_{t}\}_{t=1}^{N_T}$ to image observations, where $N_T$ is the number of frames. 
Each $\boldsymbol{M}_{t}$ consists of $\boldsymbol{\theta}_{t}$, $\boldsymbol{\beta}_{t}$, $\boldsymbol{R}_t$ and $\boldsymbol{T}_t$. 
Some previous works \cite{taylor2016efficient, mueller2019real, mueller2018ganerated, han2020megatrack} follow a similar formulation but usually optimize hand parameters $\boldsymbol{M}_{t}$ per frame, considering only the short-term temporal constraints. 
Instead, since there are no significant hand pose changes in our setting, we propose to optimize the full-sequence hand parameters $\boldsymbol{\mathcal{M}}$ at once and share the hand parameters for the whole video: $\boldsymbol{\theta}_{1} = \boldsymbol{\theta}_{2} = \dots = \boldsymbol{\theta}_{t}$, $\boldsymbol{\beta}_{1} = \boldsymbol{\beta}_{2} = \dots = \boldsymbol{\beta}_{t}$. 
Sharing hand parameters reduces the number of optimization parameters and enforces pose consistency across the whole video, which makes the hand model fitting much faster and more robust to occlusions (See Appendix C).
Specifically, we minimize the following energy function to optimize the hand parameters:
\begin{equation}
  \boldsymbol{\mathcal{M}} = \min_{\boldsymbol{\mathcal{M}}}(E_{2D} + \omega_1 E_{t} + \omega_2 E_{reg}) \text{.}
  \label{eq:total energy}
\end{equation}
The error term $E_{2D}$ evaluates the consistency between the recovered hand model and the input video:
\begin{equation}
  E_{2D} = \sum_{t}||\pi(J(\boldsymbol{M}_{t}))-J_{t}||_{2}^{2} \textbf{,}
\end{equation}
where $\mathcal{J}=\{J_t\}_{t=1}^{N_T}$ are detected 2D joints in all frames, $\pi$ is the projection operator that projects 3D hand keypoints to the image plane using the camera intrinsic matrix $\boldsymbol{K}$ initialized by image width and height.
$\boldsymbol{M}_{t}$ is the $t$-th hand model parameters. $J(\boldsymbol{M}_{t})$ is the 3D keypoints of the $t$-th hand model. 

$E_{t}$ is a term to force temporal smoothness and $E_{reg}$ is a regularization term to avoid abnormal hands:
\begin{equation}
  E_{t} = \sum_{t}||\boldsymbol{R}_t - \boldsymbol{R}_{t-1}|| + ||\boldsymbol{T}_t - \boldsymbol{T}_{t-1}|| \text{,}
\end{equation}
\begin{equation}
  E_{reg} = \sum_{t}||\boldsymbol{\mathcal{\theta}_t}||_2^2 + ||\boldsymbol{\mathcal{\beta}_t}||_2^2 \text{.}
\end{equation}  
We set $\omega_{1}$ and $\omega_{2}$ in Equation~(\ref{eq:total energy}) as 1e-4 and 5e-4 respectively in all experiments. 

Directly fitting a hand model to 2D keypoint observations is highly non-linear and sensitive to initialization. Similar to SPIN \cite{kolotouros2019spin}, we improve the fitting convergence by initializing hand model parameters for each frame with a neural network.
Specifically, we train a MANO estimation network to provide the initialization.
Then, we optimize the hand parameters in a multi-stage manner. First, we freeze the $\boldsymbol{\theta_t}$ and $\boldsymbol{\beta_t}$ and only optimize the $\boldsymbol{R}_t$ and $\boldsymbol{T}_t$ with $E_{2D}$. Then, we optimize $\boldsymbol{\mathcal{M}}$ by all the energy function in Equation~(\ref{eq:total energy}). 
During the optimization process, we use the LBFGS as the optimizer. 
For more accurate 2D keypoint detection under egocentric-view and hand-object interactions, we train a 2D keypoint detection network on our mixed dataset (See Appendix B.1).

After hand tracking,  we are able to convert the video frames to calibrated multi-view images in the hand/object-centric coordinates by the estimated $\{\boldsymbol{R}_t, \boldsymbol{T}_t\}$, and $\boldsymbol{K}$.
\subsection{Dense Reconstruction}
\label{sec:dense reconstruction}

For dense reconstruction, we choose the Signed Distance Function (SDF) as the implicit surface representation.
The SDF representation is capable of representing the high-quality object surface, from which meshes can be easily extracted by marching cubes~\cite{lorensen1987marching}.
To recover the SDF, we leverage the recent differentiable SDF rendering technique  \cite{wang2021neus}, which converts the SDF to a radiance field, renders images with volume rendering, and compares rendered images with input images to optimize the SDF.

\subsubsection{Surface representation.}
We follow \cite{wang2021neus} and represent the 3D geometry and appearance to be reconstructed as:
\label{sec:representation}
\begin{equation}
[s(\textbf{p}), c(\textbf{p}, \textbf{d})] = F(\textbf{p}, \textbf{d}) \text{,}
\end{equation}
where $F$ is an MLP network. $F$ takes 3D location $\textbf{p}$ and viewing direction $\textbf{d}$ as input, and predicts RGB color $c(\textbf{p}, \textbf{d})$ and SDF value $s(\textbf{p})$. 
Positional encoding functions are applied to $\textbf{p}$ and $\textbf{d}$ for capturing high-frequency signals.

As for rendering, for each pixel we sample a set of points along the camera ray passing through this pixel, which are denoted by $\{\textbf{p}(z)|\textbf{p}(z) = \textbf{o} + z\textbf{d},z \in [z_n,z_f]\}$ where $\textbf{o}$ is the origin of the camera, $\textbf{d}$ is the viewing direction of each image pixel, $z_n$ and $z_f$ denote the near and far bounds of the ray. 
Note that $\textbf{o}$ and per-pixel $\textbf{d}$ can be computed from hand parameters \{$\boldsymbol{R_t}$, $\boldsymbol{T_t}$\} and $\boldsymbol{K}$.
Then, the color of the pixel can be composed as
\begin{equation}
  \begin{split}
    \hat{C} &= \int_{z_n}^{z_f} \omega (z) c(\textbf{p}(z), \textbf{d})dz \text{,}
  \end{split}
  \label{eq:color rendering}
\end{equation}
where $\omega (z)$ is the weight function for accumulating colors. 
$\omega (z)$ is the product of $T(z)$ and $\rho(z)$. 
$T(z)$ measures the total transmittance accumulated from $z_n$ to $z$. $\rho(z)$ is the density function that indicates the probability of occupancy. 
We leverage the 3D hand keypoints $J(\boldsymbol{M})$ to automatically determine the $z_n$ and $z_f$, which are required to tediously manually set and tune in the previous method.
Following NeuS~\cite{wang2021neus}, an unbiased and occlusion-awareness function $\omega (z)$ is used to convert SDF values $s(\textbf{p}(z))$ to $T(z)$ and $\rho(z)$:
\begin{equation}
  T(z) = \exp(-\int_{z_n}^{z_f}\rho(z)dz) \text{,}
\end{equation}
\begin{equation}
  \rho(z) = \max \bigg (\frac{-\frac{d\Phi_h}{dz}(s(\textbf{p}(z)))}{\Phi_h(s(\textbf{p}(z)))}, 0 \bigg) \text{,}
\end{equation}
where $\Phi_h (x) = (1+e^{-hx})^{-1}$ is the Sigmoid function, $h^{-1}$ is a trainable parameter during optimization. 
In practice, the above formulations are numerically approximated using quadrature. 

\subsubsection{Camera Pose Refinement}
\label{sec:barf}
Due to the heavy occlusion between the hand and object, the camera poses (relative to the object) generated by hand tracking tend to be imprecise and noisy, which directly degrades the reconstruction quality.
To alleviate this problem, we propose simultaneously optimizing the SDF and camera poses.
By regarding camera poses as a set of optimizable parameters, they are gradually refined for lower rendering loss during the network training process.

During the simultaneous optimization, we use the coarse-to-fine strategy for more accurate camera pose refinement, following \cite{lin2021barf, park2021nerfies}.
Specifically, we weight the positional encoding for $\textbf{p}$ as $\lambda(\textbf{p}) = (\textbf{p} \dots, w_k(n_s) \cdot sin(2^k \pi \textbf{p}), w_k(n_s) \cdot cos(2^k\pi \textbf{p}) \dots)$ where $k \in [0, L-1]$. 
In our paper, we set $L$ as 6.
$n_s \in [0, N_s]$ denotes the current training step and $N_s$ is the total training steps.
$w_k$ is a weight function defined as:
\begin{equation}
  w_k(n_s) = \frac{1}{2} [1-\cos(\rm{clamp}(\frac{2n_{s}L}{N_s}-k, 0, 1) \cdot \pi)] \text{.}
\end{equation}
Starting from $n_s=0$, the positional encodings are gradually activated. When $n_s=\frac{N_{s}}{2}$, all $\lambda$ are enabled.
A similar coarse-to-fine strategy is also applied to $\textbf{d}$. Our results and ablation demonstrate that the simultaneous optimization of camera poses and radiance field significantly improves the reconstruction quality.

\subsubsection{Deformation Field}
\label{sec:deformation field}

Even though we assume the object is grasped firmly, the relative motion between the hand and object is inevitable.
The minor relative motion breaks the rigid body assumption of our multi-view reconstruction and results in artifacts in the reconstructed 3D model.
Inspired by \cite{park2021nerfies}, we equip the deformation field network $W$ to our model to compensate for the minor relative motion.
The network $W$ maps a 3D location $\textbf{p}$ from each frame to a point $\textbf{p}^{\prime}$ in a canonical space: $W: (\textbf{p}, \textbf{e}) \rightarrow \textbf{p}^{\prime}$, where $\textbf{e}$ is the per-frame temporal embedding.
With the help of the deformation field, the relative motion between the hand and object can be modeled and optimized along with the reconstruction process, which allows us to accurately recover the object geometry in the canonical space.

Specifically, the motion of each sampled point is represented by a rigid transformation in SE(3). A simple linear network is used to represent $W$, which takes $\textbf{p}$ and $\textbf{e}$ as input and outputs the 6D rotation and translation of $\textbf{p}$.

\subsubsection{Leveraging Semantics}
\label{sec:semantic-nerf}

So far, we can reconstruct the SDF of the object and the hand. Two problem remains. 
First, because the hand often occupies a large image region, uniformly sampling tends to optimize a high-quality hand to minimize the render loss and pay less attention to the object, which leads to low object reconstruction quality, particularly for small objects. 
Second, the recovered SDF contains both the hand and the object, which requires post-processing to separate the object from the hand.
We leverage semantic information to solve the above problems.

\paragraph{Semantics-guided sampling}
Since the uniform ray sampling strategy will make the network optimize less for the object part of the SDF, we use the semantic maps to guide ray sampling to focus more on the object than the hand. 
Specifically, during the training process, we gradually decrease the number of sampled rays in the hand areas and increase the number in object areas.
We sample 80\% of rays from the object region and 20\% of the hand and background region. Such a training strategy can significantly improve the sampling efficiency and reconstruction quality of the object, while maintaining the high quality of hand mesh and avoiding irregular SDF values for regions that lack supervision.

\paragraph{Semantic head}
Inspired by semantic-nerf \cite{Zhi:etal:ICCV2021}, 
we add an additional semantic head to the network $F$, which enables $F$ to predict 3D semantic labels: $[s(\textbf{p}), l(\textbf{p}), c(\textbf{p},\textbf{d})] = F(\textbf{p}, \textbf{d})$,  where $l$ is the semantic logits.
Per-ray semantic logits $\hat{L}$ can be rendered similar to color rendering Equation~(\ref{eq:color rendering}):
\begin{equation}
  \hat{L} = \int_{z_n}^{z_f} \omega(z) l(\textbf{p}(z), \textbf{d})dz \text{.}
\end{equation}
The 3D semantic prediction is rendered and supervised by the 2D semantic maps.

\subsubsection{Training Loss.}
We utilize multiple loss terms to optimize our hand-object SDF. 
To compute the loss, we sample $N_r$ rays on each image and $N_p$ points along each ray.
The total loss function is defined as:
\begin{equation}
  L = L_{c} + \lambda_{m} L_{m} + \lambda_{e} L_{e} + \lambda_{l} L_{l} + \lambda_{d} L_{d} \text{.}
\end{equation}
$L_c$ is a color rendering loss between rendered pixel colors $\hat{C}$ and true values $C$ of sampled rays in training images:
\begin{equation}
    L_c = \frac{1}{N_r} \sum_i ||C_i - \hat{C}_i||_2^2 \text{.}
\end{equation} %
The Eikonal term \cite{Gropp2020ImplicitGR} $L_{e}$ and mask term $L_m$ are used to regularize the SDF:
\begin{equation}
  L_{e} = \frac{1}{N_r N_p}\sum_{i,j}(|\nabla s(\textbf{p}_{i,j})|-1)^2 \text{,}
\end{equation}
\begin{equation}
  L_m = \rm{CE}(M_i, \hat{O}_i) \text{,}
\end{equation}
where the CE denotes the cross-entropy loss. $M_i$ is the mask value and $\hat{O}_i$ is the sum of weight $\omega(t)$ along the ray. 
We set $\lambda_{e}$ as 0.1, same with NeuS, but use a larger $\lambda_{m}=5.0$ to remove the influence of inconsistent motion between the foreground and background. 
$L_{l}$ is the semantic logits rendering loss, which uses a multi-class cross-entropy loss to optimize the semantic branch:
\begin{equation}
    L_l = \rm{CE}(L_i, \hat{L}_i) \text{,}
\end{equation}
where $L_i$ denotes the ground-truth semantic labels. $\lambda_{l}$ is set as 0.1 in experiments.
$L_{d}$ is a deformation regularization term proposed in \cite{tretschk2021nonrigid}:
\begin{equation}
    L_{d} = \frac{1}{N_r N_p} \sum_{i,j} \rho(\textbf{p}_{i,j}) \cdot |\rm{div} (\textbf{p}_{i,j}^{\prime}-\textbf{p}_{i,j})|^2,
\end{equation}
where $\lambda_{d}$ is set as 10.0 empirically.

\subsubsection{Post-processing}
Given the learned neural implicit field $F$, the 3D geometry of hand and object can be reconstructed from the explicit SDF volume by querying $F$.
Then, the predicted 3D semantic labels of each vertex are used to separate the hand and the object. 
Finally, Poisson Reconstruction~\cite{kazhdan2006poisson} is used to fill the holes on the reconstructed object mesh caused by occlusion from the hand.
\section{EXPERIMENTS}

\subsection{Implementation}
\label{sec:implementation}
We use a single NVIDIA TITAN RTX GPU to run the whole reconstruction pipeline. For a one-minute 4K video, the hand motion capture step takes 15 minutes, including 8 minutes for hand 2D keypoint estimation and 7 minutes for MANO fitting. Then, the video is downsampled from 30fps to 6fps to perform the segmentation and reconstruction. 
The segmentation networks take 6 minutes. The neural SDF reconstruction requires 30 hours to converge. Our implementation is based on the PyTorch~\cite{paszke2019pytorch} library.

\subsection{Dataset}
\label{sec:dataset}
Since no existing dataset can satisfy our setting, we collect a new 3D object reconstruction dataset, called Hand-held Object Dataset (HOD). 
The HOD dataset contains 35 objects, which is divided into two subsets named \textit{Sculptures} and \textit{Daily Objects}. The \textit{Sculptures} has five human sculptures with complex geometries and pure white textures. The \textit{Daily Objects} consists of 30 daily objects with various shapes and appearances. 
All of the \textit{Sculptures} and nine of the \textit{Daily Objects} are paired with high-fidelity scanned meshes as ground truth geometries for evaluation.

\begin{figure}[htb]
\begin{center}
  \includegraphics[width=1.0\linewidth, trim=0 100 0 40]{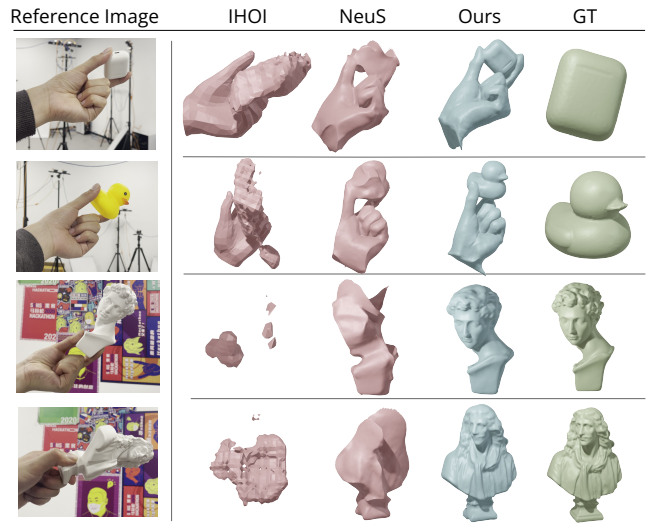}
  \caption{
    {\textbf{Qualitative results of hand-held object reconstruction.}}
    \textnormal{We show the joint reconstruction of the hand and the object for the top two examples, and the segmented object meshes for the bottom two examples.
    IHOI \cite{ye2022hand} is a learning-based single-view reconstruction method with the reference frame as input. NeuS \cite{wang2021neus} uses the same video frames and camera poses as ours as input. \textit{The full figure with more examples and compared methods is in Appendix D.8.}}
  }
\label{fig:comparison}
\end{center}
\vspace{-0.3 cm}
\end{figure}

\subsection{Evaluations}
\label{sec:evaluations}

\subsubsection{Metrics and Baseline}
We use Chamfer Distance (CD) as the metric for quantitative evaluation. Since the reconstructed and the ground-truth mesh lay in different coordinate systems, we normalize each mesh to unit size and register the reconstructed mesh to the ground-truth mesh using the point-to-point iterative closest point (ICP) algorithm.

We compare our method against the state-of-the-art multi-view reconstruction methods and learning-based hand-object reconstruction methods. 
We use NeuS \cite{wang2021neus} as the multi-view baseline method which our method is built upon. The original NeuS uses COLMAP to obtain camera poses which does not work in our setting, so we feed the camera poses estimated from our hand tracking stage to NeuS.
To the best of our knowledge, there is no previous learning-based method that reconstructs hand-object interactions from a monocular video. We then compare our method against ObMan \cite{hasson19_obman}, GF \cite{karunratanakul2020grasping} and IHOI \cite{ye2022hand}, all of which reconstruct hand-object interactions from a single image using deep neural networks.
Since learning-based methods take a single image as input, we evaluate them on each frame of the video and report the average accuracy.

\subsubsection{Quantitative evaluation}

We provide quantitative results for both \textit{Sculptures} and \textit{Daily Objects} subsets in Table~\ref{tab:chamfer}.
The results show that our method outperforms ObMan, GF, and IHOI with much lower CD using the same hand-held object capture video as input. As learning-based methods heavily rely on the learned object shape prior to predicting the 3D object geometry, they don't work well for objects beyond the training dataset.
NeuS obtains more reasonable results than the single-view prediction methods but still fails to reconstruct accurate shapes as the original NeuS cannot handle the imprecise pose estimation and relative hand-object motion in our problem. 
Our approach outperforms the baseline method by a large margin, which validates the effectiveness of the three proposed components.

\begin{table}[tb]
    \caption{Quantitative results of object reconstruction.
        \textnormal{The metric is the Chamfer Distance.}
        }
    \label{tab:chamfer}
    
    \centering
    \begin{tabular}{c||c|c|c||c||c}
    \Xhline{3\arrayrulewidth}
        ID & ObMan & GF & IHOI & NeuS & Ours \\ \hline
        Orange      & 3.426 & 158.735 & 12.055 & 3.963   & \textbf{0.304} \\
        Plastic Box & 1.969 & 24.645  & 3.866  & 1.515   & \textbf{0.433} \\
        Rubber Duck & 3.984 & 59.713  & 7.676  & 2.035   & \textbf{0.521} \\
        Robot       & 2.250 & 27.589  & 4.703  & 1.354   & \textbf{0.207} \\
        Cat         & 8.808 & 56.905  & 15.334 & 11.192  & \textbf{0.225} \\
        AirPods     & 0.721 & 24.561  & 6.978  & 1.081   & \textbf{0.083} \\
        Bottle      & 0.711 & 271.218 & 3.995  & 0.875   & \textbf{0.293} \\
        Case        & 2.438 & 28.630  & 7.065  & 0.929   & \textbf{0.242} \\
        Pingpong    & 8.588 & 112.620 & 12.759 & 4.381   & \textbf{0.408} \\ \hline

        Apollo      & 3.387 & 45.322  & 4.255  & 8.724   & \textbf{0.164} \\
        David       & 1.806 & 40.947  & 3.652  & 3.348   & \textbf{0.191} \\
        Giuliano    & 1.448 & 20.794  & 4.139  & 0.873   & \textbf{0.094} \\
        Marseille   & 3.012 & 60.904  & 4.698  & 1.918   & \textbf{0.181} \\
        Moliere     & 1.737 & 28.963  & 4.054  & 5.290   & \textbf{0.145} \\ \hline
        mean        & 3.163 & 68.682  & 6.802  & 3.391   & \textbf{0.249} \\
    \Xhline{3\arrayrulewidth}
    \end{tabular}

\end{table}

\subsubsection{Qualitative evaluation}

We show the qualitative results of hand-held object reconstruction in Figure~\ref{fig:comparison} (full figure in Appendix D.8).
NeuS can generate the approximate shape of the target object but with many sharp artifacts. This is mainly due to the imprecise camera poses and not considering the small relative motion between the hand and object.
The learning-based method IHOI cannot reconstruct unseen objects and generate invalid mesh for the given image. 
In contrast, our reconstructed meshes are more fidelity than IHOI and NeuS.

\subsubsection{Comparison with static object capture}
To demonstrate the reconstruction ability of our proposed hand-held object capture pipeline, we conduct additional comparative experiments with COL-MAP \cite{ schoenberger2016mvs} and NeuS \cite{wang2021neus} in the \textit{static object capture} setting.
Specifically, we capture an extra one-minute 4K video for each object by placing the object on a flat table and slowly moving the camera surrounding the object to capture the video. 
Textured papers are put beneath the object to provide extra textures for more accurate camera pose estimation of SfM.

\begin{figure}[tb]
\begin{center}
  \includegraphics[width=1.0\linewidth, trim=0 200 0 40]{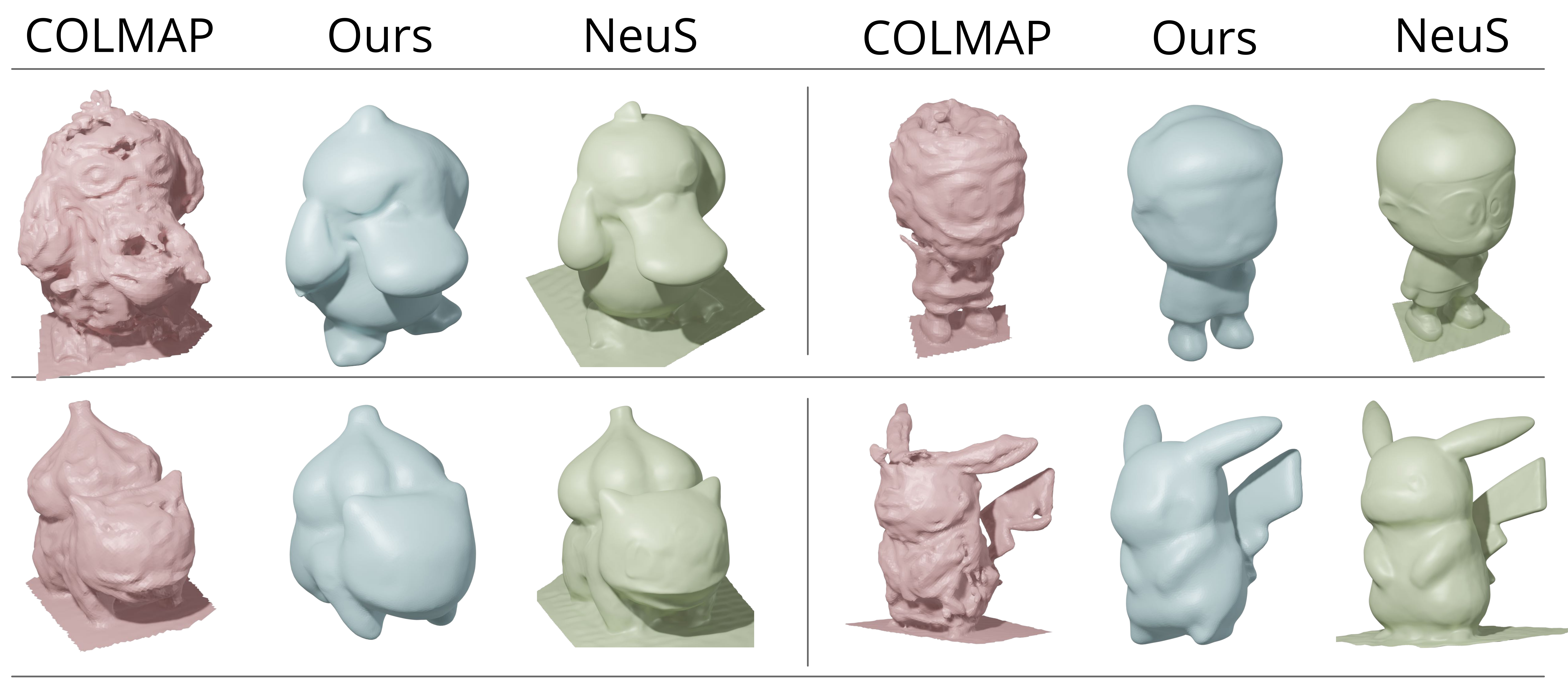}
  \caption{{\textbf{Comparison with static object capture.}}
  \textnormal{Note that the inputs to COLMAP and NeuS are videos of static object capture by putting the object on a table, while our inputs are videos of hand-held object capture by moving the object in front of the camera.}
  }
\label{fig:colmap}
\end{center}
\end{figure}

The qualitative and quantitative results are shown in Figure \ref{fig:colmap} and Appendix Table 3. 
Even with hand-held capturing, our method obtains higher object reconstruction quality than COLMAP with a static capture setting. 
However, our method cannot recover as many fine details as NeuS.
The reason is that the input of NeuS has accurate camera poses and little occlusion. In contrast, our input videos contain heavy hand-object occlusions, and it is hard to recover accurate camera poses.
Nevertheless, this comparison shows that our pipeline is able to achieve competitive reconstruction quality that is close to static object capture, while the data capture process is more user-friendly.

\subsubsection{Robustness}
We demonstrate the robustness of our method with different grasping gestures and different motion speeds in Appendix D.1 and D.2.

\subsection{Ablation Study}
\label{sec:ablation}

\begin{figure}[tb]
\begin{center}
  \includegraphics[width=1.0\linewidth, trim=0 100 0 40]{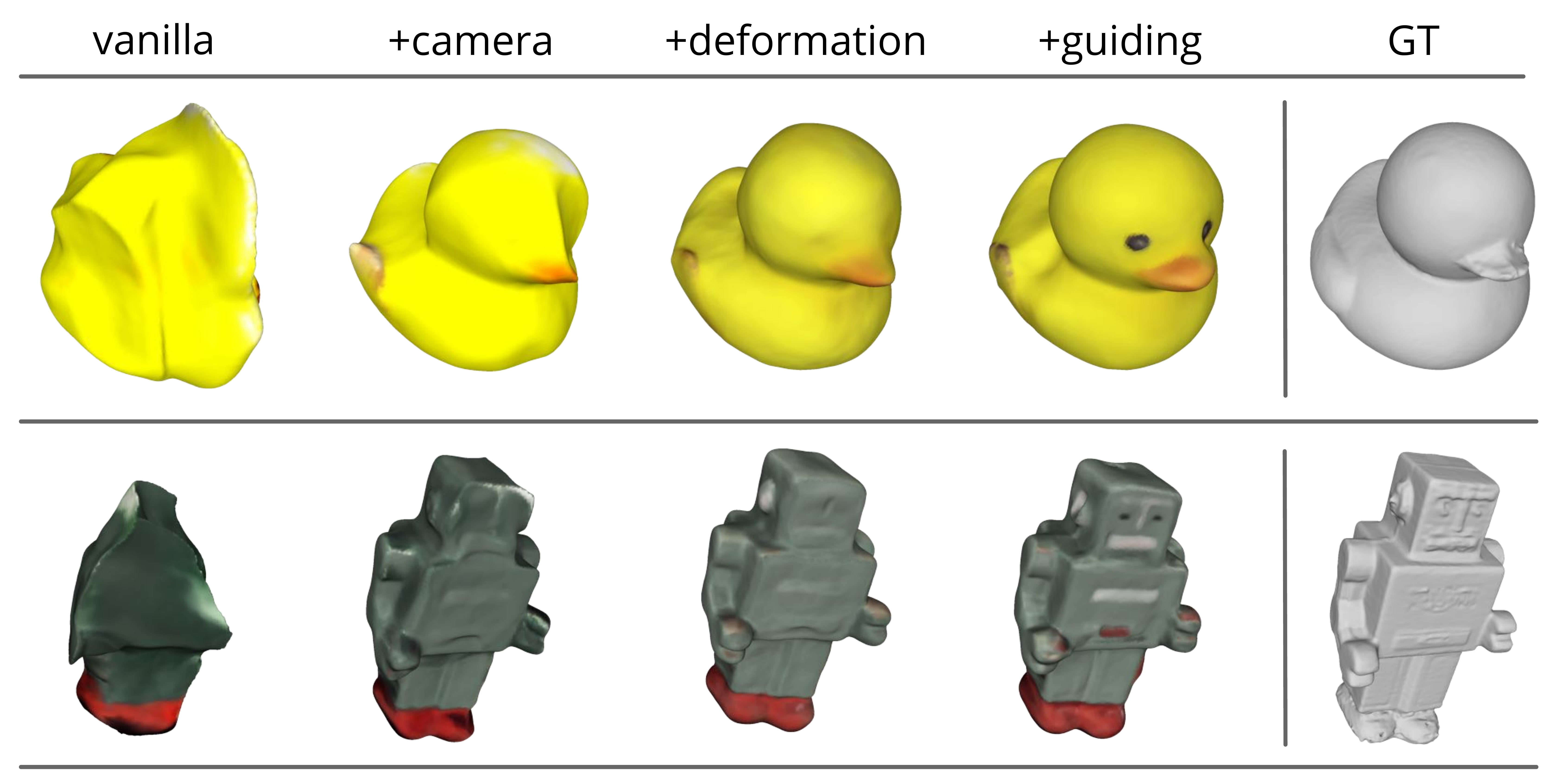}
  \caption{{\textbf{Qualitative results of ablation study.}}
       \textnormal{\textit{vanilla} indicates the mesh reconstructed with original NeuS, \textit{+camera} indicates the camera refinement introduced in Section~\ref{sec:barf}, \textit{+deformation} indicates the deformation field introduced in Section~\ref{sec:deformation field}, and \textit{+guiding} indicates the semantics-guided sampling introduced in Section~\ref{sec:semantic-nerf}.}
  }
\label{fig:ablation}
\end{center}
\vspace{-0.4 cm}
\end{figure}

In this paper, we propose several modifications to NeuS to improve the reconstruction quality. 
We demonstrate their effectiveness by sequentially adding them to the vanilla NeuS model.
As shown in Figure~\ref{fig:ablation}, the vanilla NeuS suffers from incorrect camera poses due to imprecise hand pose tracking. By simultaneously optimizing the camera poses, the optimized SDF improves but still contains sharp artifacts. By adding the deformation field, the network $F$ can capture relative motion between the hand and object, which removes the irregular sharp edges. Finally, the semantics-guided sampling makes the network more focused on the object instead of the hand, enabling more object details to be reconstructed. 
Please see Appendix D.6 for quantitative results.
\section{Limitations and Future Work}

In this paper, we showed that it was possible to reconstruct a moving hand-held object which may be textureless from a monocular video, without using any shape prior and training data of the object. 
There are some limitations of this work.
First, Poisson Reconstruction is used as post-processing for hole filling, which may not be suitable for large holes or thin structures. More sophisticated learning-based shape completion methods could be used.
Second, depth-based in-hand capture allows the user to see the reconstruction in real-time, while our method takes hours to reconstruct the object. Leveraging recent techniques \cite{mueller2022instant, TensoRF} to accelerate our method is left as future work.
Finally, our method assumes a fixed camera, a single grasping gesture, and small relative motion between hands and objects. Extending our method to relax these constraints would be an interesting direction to explore.

\begin{acks}
  The authors would like to acknowledge the support from ZJU-SenseTime 3D Vision Lab and the Open Project Program of the State Key Lab of CAD\&CG (No.A2213), Zhejiang University.
\end{acks}

\bibliographystyle{ACM-Reference-Format}
\bibliography{bibliography}

\clearpage
\appendix

\section{Dataset Details}
\label{sec:dataset details}

\begin{figure}[htb]
\begin{center}
  \includegraphics[width=1.0\linewidth]{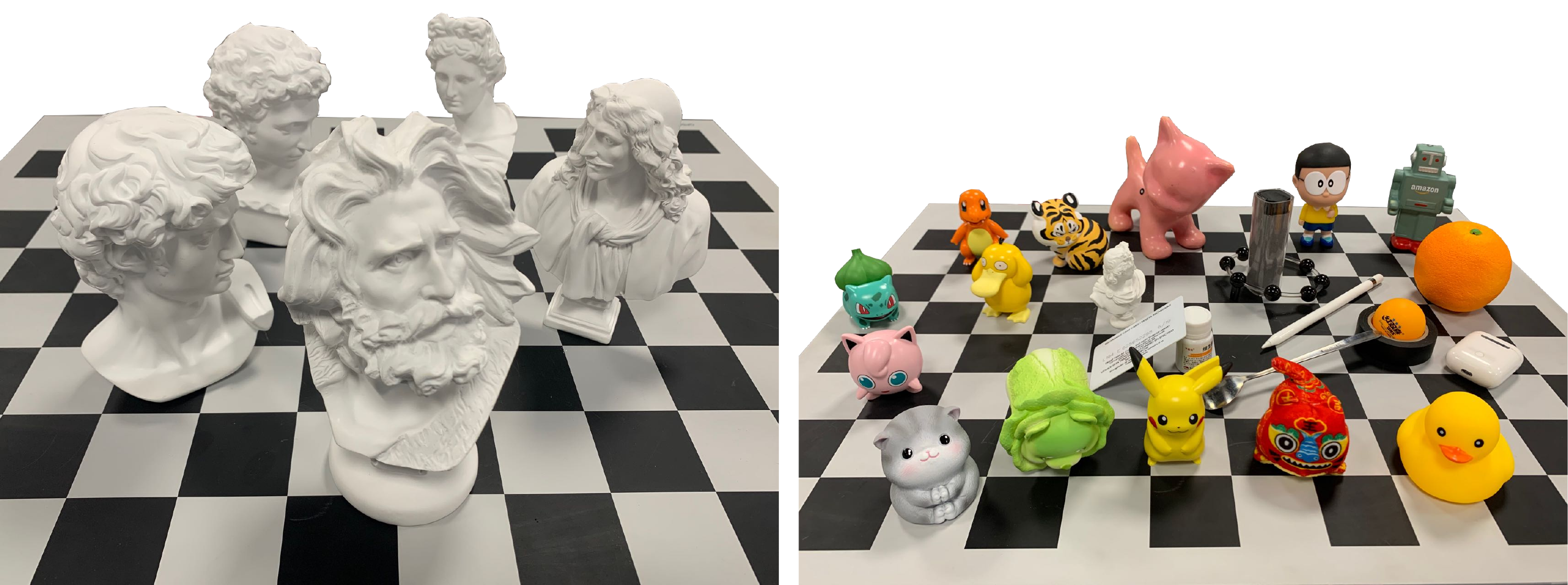}
  \caption{{\textbf{HOD dataset.}}
    \textnormal{The left figure shows five textureless \textit{Sculptures} with complex geometry details. The right figure shows some examples of the \textit{Daily Objects} in the HOD dataset, which contains various shapes and textures. }
  }
\label{fig: dataset}
\end{center}
\end{figure}

An overview of the dataset objects is given in Figure \ref{fig: dataset}. 
To inspire future hand-held object reconstruction works, each object is captured with extra videos for two additional settings: 
1) \textbf{Hand-held Object Reconstruction with Different Gestures}. 
The classic static object scanning recovers the bottom surface of the object by scanning the object a second time with a different object placement.
Likewise, for hand-held object reconstruction, it should be possible to reconstruct the occluded region by changing the position of contact points. 
We then capture an additional video for each object with a different grasping gesture. 
The main challenge of solving this setting is building the accurate correspondence between two partial object meshes and fusing two slightly different meshes.
2) \textbf{Hand-held Object Reconstruction with Large Hand-object Relative Motion}.
Unlike our work, a more user-friendly setting should allow significant relative motion between the hand and object. Thus, we capture a video with the large hand-object relative motion for each object.
To solve this setting, future work should be able to infer the object pose by human manipulation and fuse the temporal observations to a complete object mesh.

There are several datasets of hand-object interactions. HO3D \cite{hampali2020honnotate} and ContactPose \cite{Brahmbhatt_2020_ECCV} are the representatives. Compared to them, our dataset is more suitable for the task of hand-held object reconstruction for several reasons:
First, both ContactPose and HO3D use multiple cameras while each camera only sees a part of the object, which makes HO3D and ContactPose not suitable for the monocular reconstruction setting.
Second, compared with HO3D, the objects in our dataset are firmly grasped by users, while HO3D allows significant relative motion between the hand and object.
Third, compared with ContactPose, our dataset has more realistic objects. ContactPose uses 3D-printed replica objects with no texture. Our dataset contains objects from daily-life scenarios with various shapes and appearances.

\section{Details of Auxiliary Network Training}
\subsection{Hand 2D/MANO network training}
\label{sec:appendix 2dmano}
The input video of our method is captured in the egocentric view and contains strong occlusions due to the hand-object interactions. However, off-the-shelf 2D hand keypoints networks \cite{simon2017hand} and MANO estimation networks \cite{lin2021end-to-end, Kulon_2020_CVPR} are mostly trained for third-person perspective and no hand-object interactions. 
To solve this problem and improve the hand tracking performance, we propose to train new hand networks with a mixture of egocentric-view and hand-object interaction datasets.
Specifically, we use FreiHand \cite{Freihand2019}, InterHand \cite{Moon_2020_ECCV_InterHand2.6M}, MTC \cite{xiang2019monocular}, RHD \cite{zb2017hand}, Ego3D \cite{Lin_2021_WACV} and HO3D \cite{hampali2020honnotate} to improve the hand tracking performance. For the network architecture, we use ResNet50 as the 2D hand keypoints estimation backbone network, and HMR \cite{kanazawaHMR18} as the MANO estimation backbone network.

\subsection{Hand segmentation network training}
\label{sec:appendix segmentation}
Since none of the current hand segmentation networks performs well on egocentric-view and hand-object interaction scenarios, we collect and annotate a new hand segmentation dataset.
The dataset contains 3000 egocentric hand-object interaction images with manually labeled hand segmentation labels. 
For the network architecture, we choose the DeepLabv3+ \cite{deeplabv3plus2018} as our backbone network.

\section{Hand Tracking with Parameters Sharing}
\label{sec:appendix hand}
\begin{figure}[htb]
\begin{center}
  \includegraphics[width=1.0\linewidth]{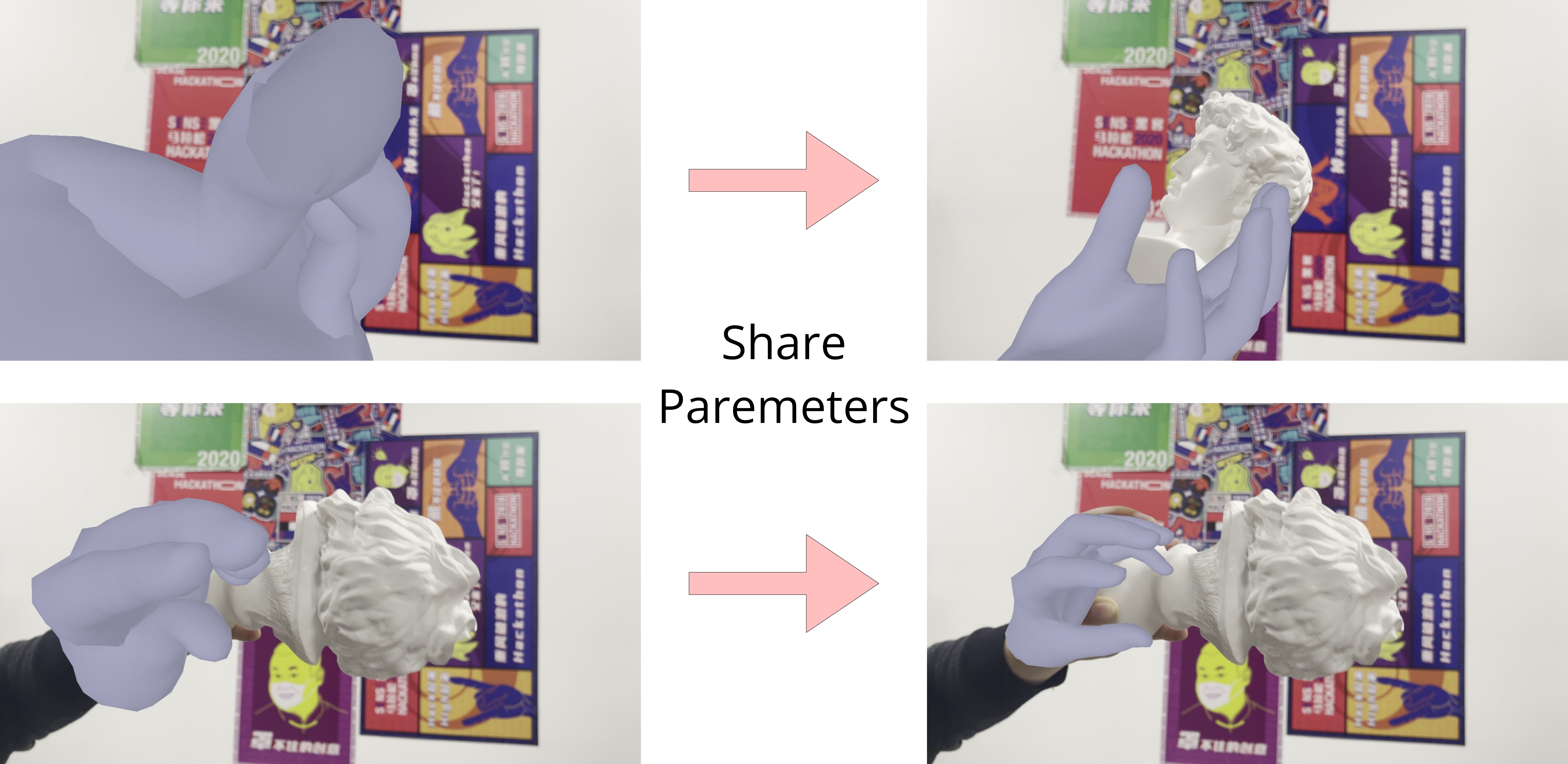}
  \caption{{\textbf{Effectiveness of parameters sharing during hand tracking.}}
    \textnormal{We compare our hand tracking system with and without sharing hand pose ($\boldsymbol{\theta}$) and shape ($\boldsymbol{\beta}$) parameters. 
    Without sharing hand parameters, the hand tracking module easily tracks wrong hand poses due to heavy hand-object occlusions. 
    In contrast, by sharing hand parameters across the entire video, the tracking module is able to get the correct hand poses even in extreme cases.}
    }
\label{fig: hand tracking}
\end{center}
\end{figure}

Here we compare our hand tracking method with and without parameters sharing. 
As mentioned in the Section 4.1, sharing parameters uses the same pose ($\boldsymbol{\theta}$) and shape ($\boldsymbol{\beta}$) parameters for each frame.
Such parameters sharing manner makes the hand tracking process much faster due to fewer optimization parameters and easier convergence.
According to our experiments on 1800 frames of video, the fitting process takes about 4600 seconds without sharing the hand parameters. 
However, the hand fitting process only takes about 420 seconds when the hand parameters are shared.

Another benefit of sharing hand parameters is the tracking robustness under heavy hand-object occlusions.
As shown in Figure \ref{fig: hand tracking}, without sharing hand parameters, the hand tracking method is sensitive to keypoints outliers and often fails to produce correct hand poses due to the heavy occlusion. However, by sharing hand parameters across the whole sequence, the tracking process can use other frames to correct wrong estimations, resulting in better tracking results.

\section{Experiments for dense reconstruction}

\subsection{Results for different motion speeds}
\label{sec:appendix speed}

\begin{figure}[htb]
\begin{center}
  \includegraphics[width=1.0\linewidth]{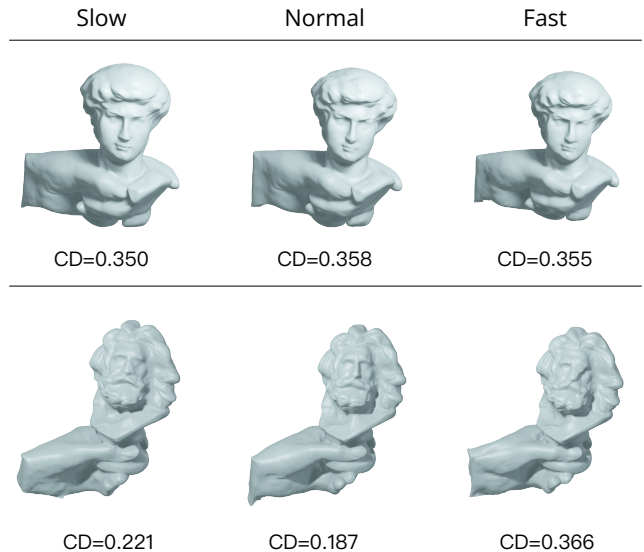}
  \caption{{\textbf{Experiments of different moving speeds.}}
    \textnormal{We do experiments to validate the robustness of our method at different grasping motion speeds. Our method is able to generate high-quality 3D geometry for the \textit{Slow} and \textit{Normal}, and only cause small artifacts for the \textit{Fast}. The `CD' refers to the Chamfer Distance of object reconstruction.}
  }
\label{fig: speed}
\end{center}
\end{figure}

\begin{figure*}[htb]
\begin{center}
  \includegraphics[width=1.0\linewidth]{figures/gesture_color.pdf}
  \caption{
    {\textbf{Experiments of different grasping gestures.}}
    \textnormal{We conduct experiments to verify whether grasping gestures affect the final reconstruction results. When the grasping gesture covers only a small part of the object, both joint hand-object reconstruction and final object-only results are high-fidelity. Our method can still generate high-quality joint hand-object reconstruction results when grasping gestures cover a large portion of the object. However, the limitations of our post-processing make it challenging to fill in holes under heavy occlusions. The `CD' refers to the Chamfer Distance of object reconstruction.}
  }
\label{fig: gesture}
\end{center}
\end{figure*}

We validate our reconstruction method at different motion speeds. We collect videos at three different speeds with the same grasping gesture. \textit{Normal} is the normal speed at which humans daily interact with objects. \textit{Slow} and \textit{Fast} are half the movement speed and twice the motion speed of \textit{Normal}.
The results of the joint hand-object reconstruction are shown in the figure \ref{fig: speed}.
It turns out that our algorithm works well for both the \textit{Slow} and \textit{Normal} speeds. For faster motion, our reconstruction results may cause some artifacts, but are still correct for most regions.

\subsection{Results for different grasping gestures}
\label{sec:appendix gestures}

We also validate our reconstruction algorithm with different grasping gestures. As shown in Figure \ref{fig: gesture}, different grasping gestures lead to varying levels of occlusions. 
No matter how the hand holds the object, our method is capable of generating high-fidelity joint hand-object reconstruction meshes since we do not use any gesture-related prior or assume specific grasping types.
However, the object-only reconstruction contains obvious artifacts when heavy occlusions happen.
The reason is that we use Poisson Reconstruction as post-processing for hole filling. When a large part of the object is occluded or covered by the hand, the Poisson Reconstruction cannot fill the correct surface for the missing part, resulting in artifacts, see \textit{Strong Occlusion} row in Figure \ref{fig: gesture}. In conclusion, gestures are irrelevant for joint hand-object reconstruction. However, for object-only reconstruction, weak occlusion gestures make it easier for Poisson Reconstruction to fill in the holes.

\begin{table}[htb]
    \caption{\textbf{Quantitative comparison with GT-selected meshes.}
        \textnormal{The metric is the Chamfer Distance between the reconstructed and the ground-truth mesh.}
        }
    \label{tab: chamfer best}
    \centering
    \begin{tabular}{c||c|c|c||c}
    \Xhline{3\arrayrulewidth}
        ID & ObMan & GF & IHOI & Ours \\ \hline
        Orange      & 0.814 & 1.080 & 1.936   & \textbf{0.304} \\
        Plastic Box & \textbf{0.282} & 0.605  & 0.575    & 0.433 \\
        Rubber Duck & 1.926 & 2.504  & 1.794    & \textbf{0.521} \\
        Robot       & 1.028 & 1.055  & 1.010    & \textbf{0.207} \\
        Cat         & 4.891 & 6.128  & 3.667   & \textbf{0.225} \\
        AirPods     & 0.314 & 0.416  & 0.377    & \textbf{0.083} \\
        Bottle      & \textbf{0.245} & 0.408 & 0.303    & 0.293 \\
        Case        & 0.641 & 0.957  & 0.757    & \textbf{0.242} \\
        Pingpong    & 1.869 & 5.286 & 2.567   & \textbf{0.408} \\ \hline

        Apollo      & 0.784 & 1.371  & 0.788    & \textbf{0.164} \\
        David       & 0.849 & 1.262  & 0.650    & \textbf{0.191} \\
        Giuliano    & 0.481 & 0.589  & 0.525    & \textbf{0.094} \\
        Marseille   & 1.217 & 1.175  & 1.118    & \textbf{0.181} \\
        Moliere     & 0.786 & 0.800  & 0.597    & \textbf{0.145} \\ \hline
        mean        & 1.152 & 1.688  & 1.190    & \textbf{0.249} \\
    \Xhline{3\arrayrulewidth}
    \end{tabular}

\end{table}

\begin{table}[tb]
    \caption{
    \textbf{Quantitative comparison with temporal smoothed meshes.} 
    \textnormal{The metric is the Chamfer Distance between the reconstructed and the ground-truth mesh. }
    }
    \label{tab:remporal}
    
    \centering
    \begin{tabular}{c||c|c|c|c}
    \Xhline{3\arrayrulewidth}
        ID & ObMan & GF & IHOI & Ours \\ \hline
        Orange & 0.552 & 1.454 & 2.314 & \textbf{0.304} \\
        Plastic & 7.325 & 0.761 & 1.208 & \textbf{0.433} \\
        Rubber Duck & 3.330 & 2.662 & 2.390 & \textbf{0.521} \\
        Robot & 1.862 & 1.007 & 2.236 & \textbf{0.207} \\
        Cat & 10.605 & 5.900 & 6.226 & \textbf{0.225} \\
        AirPods & 0.210 & 0.338 & 0.432 & \textbf{0.083} \\
        Bottle & 0.263 & 1.258 & \textbf{0.281} & 0.293 \\
        Case & 1.586 & 1.972 & 1.967 & \textbf{0.242} \\
        Pingpong & 4.262 & 4.471 & 1.756 & \textbf{0.408} \\ \hline
        mean & 3.332 & 2.203 & 2.090 & \textbf{0.302} \\
    \Xhline{3\arrayrulewidth}
    \end{tabular}
    
\end{table}

\subsection{Comparison with GT-selected meshes}
\label{sec:gt-select}

To demonstrate the quality of our reconstructed meshes, we also compare them with the meshes of learning-based methods selected by the ground-truth. 
Specifically, we run the learning-based hand-object reconstruction method per frame, align them with the ground-truth by ICP, calculate the Chamfer Distance, and select the most similar mesh with the ground-truth. 
The table \ref{tab: chamfer best} shows that even compared with the GT-selected meshes of learning-based approaches, our method still gets a lower Chamfer Distance by a large margin.  
The only exception is the \textit{Plastic Box} and \textit{Bottle} compared with ObMan. 
We argue that the reason is those two objects are cylinder-like. 
According to our experiments, ObMan is prone to generate cylinder-like geometries. For this reason, ObMan gains better Chamfer Distance results for the two instances. 
However, ObMan heavily relies on learned object prior and often generates cylinder-like shapes for every unseen object, thus cannot generate correct geometry for other instances. In contrast, our method can generalize to various types of objects and gain much better performance, especially for those with complex geometries or low texture information.

\subsection{Comparison with temporal smoothed meshes}
A simple strategy to boost the performance of learning-based methods is leveraging temporal coherence. 
We apply temporal smoothing to learning-based methods and compare them with the proposed pipeline. 
Specifically, we run learning-based methods to reconstruct the object mesh for each frame. The reconstructed object meshes are then aligned with the ground-truth mesh by ICP and voxelized to 3D occupancy volumes. 
We average the 3D occupancy volume across the whole temporal sequence, then set a threshold and use marching cubes to extract the final mesh.
The quantitative results are shown in Table \ref{tab:remporal}. 
By leveraging temporal coherence, the learning-based methods are able to fuse multi-frame observations and get better performance.
Compared with learning-based methods with temporal coherence, our method still gets a better reconstruction quality and lower Chamfer Distance by a large margin. The only exception is the \textit{Bottle}, which has a cylinder-like shape whose shape prior is easier to learn by learning-based methods (Discussed in Section \ref{sec:gt-select}).

\begin{figure}[htb]
  \begin{center}
    \includegraphics[width=1.0\linewidth]{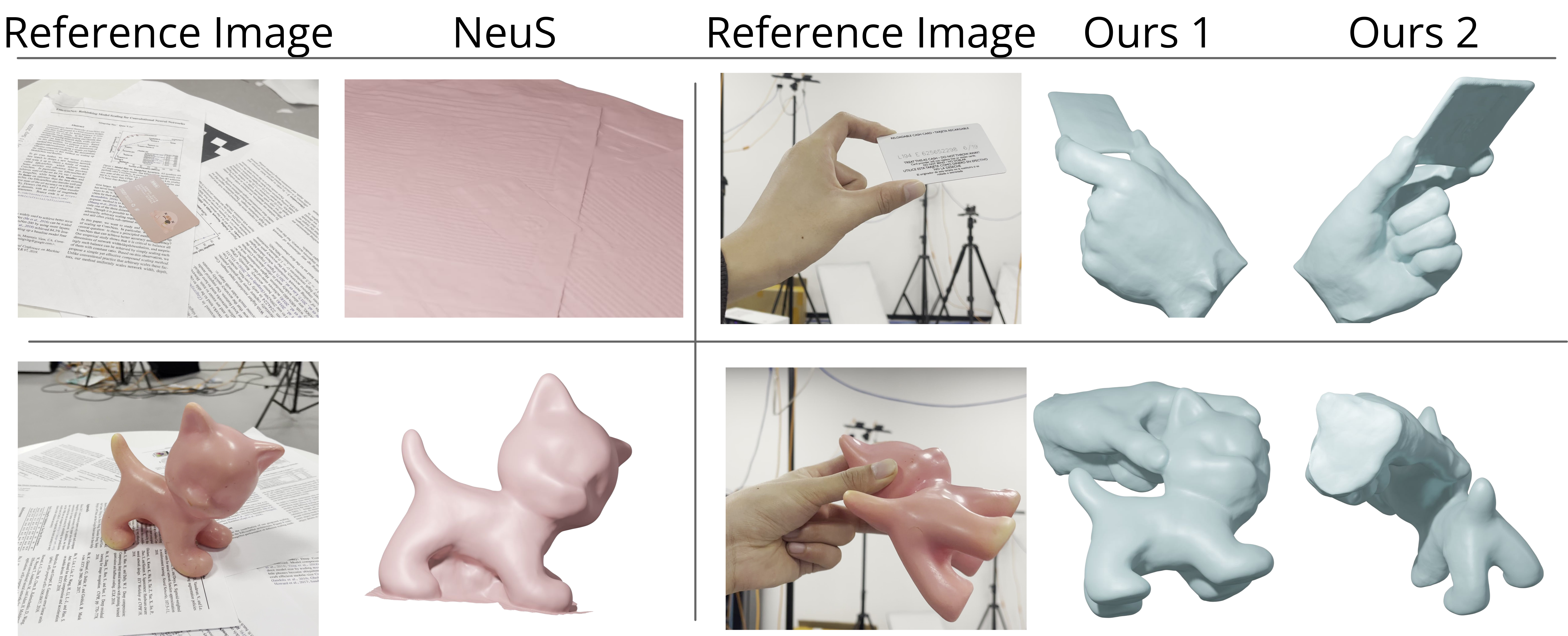}
    \caption{
    {\textbf{Comparison with static object capture.}}
    \textnormal{The proposed setting is able to reconstruct objects that cannot stand vertically, or the bottom surfaces are heavily occluded.}
    }
  \label{fig: benefits}
  \end{center}
\end{figure}
\begin{table}[tb]
    \caption{
    \textbf{Quantitative comparison with static object capture.} 
    \textnormal{The metric is the Chamfer Distance between the reconstructed and the ground-truth mesh. }}
    \label{tab:static-comparison}

    \centering
    \begin{tabular}{c||c|c|c}
    \Xhline{3\arrayrulewidth}
        ID & COLMAP  & NeuS & Ours  \\ \hline
        Orange & 0.260 & \textbf{0.232} & 0.304  \\
        Plastic & 0.350 & \textbf{0.062} & 0.433  \\
        Rubber Duck & 1.917 & \textbf{0.326} & 0.521  \\
        Robot & \textbf{0.118} & 0.123 & 0.207  \\
        Cat & 1.049 & \textbf{0.173} & 0.225  \\
        AirPods & 0.895 & 0.228 & \textbf{0.083}  \\
        Bottle & 0.043 & \textbf{0.032} & 0.293  \\
        Case & 0.483 & \textbf{0.172} & 0.242  \\
        Pingpong & 1.081 & \textbf{0.037} & 0.408  \\ \hline
        mean & 0.688 & \textbf{0.154} & 0.302 \\
    \Xhline{3\arrayrulewidth}
    \end{tabular}
    
\end{table}

\begin{table}[tb]
    \caption{\textbf{Quantitative results of ablation study.} 
    \textnormal{The metric is the PSNR of the target object region. }
    }
    \label{tab:psnr}

    \centering
    \begin{tabular}{c||c|c|c|c}
    \Xhline{3\arrayrulewidth}
        ID & vanilla  & +camera & +deformation & +guiding \\ \hline
        Orange & 20.082 & 21.897 & 22.576 & \textbf{28.756} \\
        Plastic & 17.156 & 18.260 & 18.989 & \textbf{25.215} \\
        Rubber Duck & 18.453 & 19.474 & 21.307 & \textbf{28.903} \\
        Robot & 17.083 & 18.522 & 19.506 & \textbf{24.857} \\
        Cat & 17.797 & 20.456 & 20.490 & \textbf{26.456} \\
        AirPods & 17.945 & 20.300 & 21.819 & \textbf{28.427} \\
        Bottle & 17.804 & 18.655 & 19.123 & \textbf{25.804} \\
        Case & 19.323 & 20.649 & 21.096 & \textbf{28.468} \\
        Pingpong & 17.277 & 18.956 & 19.627 & \textbf{28.508} \\ \hline
        mean & 18.102 & 19.685 & 20.504 & \textbf{27.266} \\
    \Xhline{3\arrayrulewidth}
    \end{tabular}
    
\end{table}

\subsection{Comparison with static object capture}
In this paper, we propose a novel object reconstruction manner by grasping and moving the object in front of the camera. Compared with  classic static object capture, the proposed method has several potential benefits:
First, our method is able to reconstruct objects that cannot be easily placed on the floor or table. 
For example, the \textit{card} from our HOD dataset is impossible to stand vertically on the floor.
As shown in the first row of Figure \ref{fig: benefits}, NeuS with static object capture cannot reconstruct the card and outputs a plane for the whole table. However, our method is able to reconstruct the \textit{card} by grasping it in hand. 
Second, static object capture requires placing the object on a plane, which makes it hard to reconstruct the bottom surface.
As shown in the second row of Figure \ref{fig: benefits}, static object capture cannot reconstruct the bottom part of the \textit{cat} due to occlusion, while our setting allows users to scan the object more freely (as long as the object can be grasped).

\subsection{Effectiveness of the three proposed modules}
\label{sec:appendix ablation}

We evaluate the PSNR metric in Table \ref{tab:psnr} as a supplement of Section 5.4. The higher PSNR value indicates a better novel view quality and mesh details. According to the table, the three solutions show their ability to solve the three main issues that degrade the reconstruction quality.

\subsection{Effectiveness of camera pose initialization}
\begin{figure}[htb]
  \begin{center}
    \setlength{\abovecaptionskip}{0.1cm}
    \includegraphics[width=1.0\linewidth]{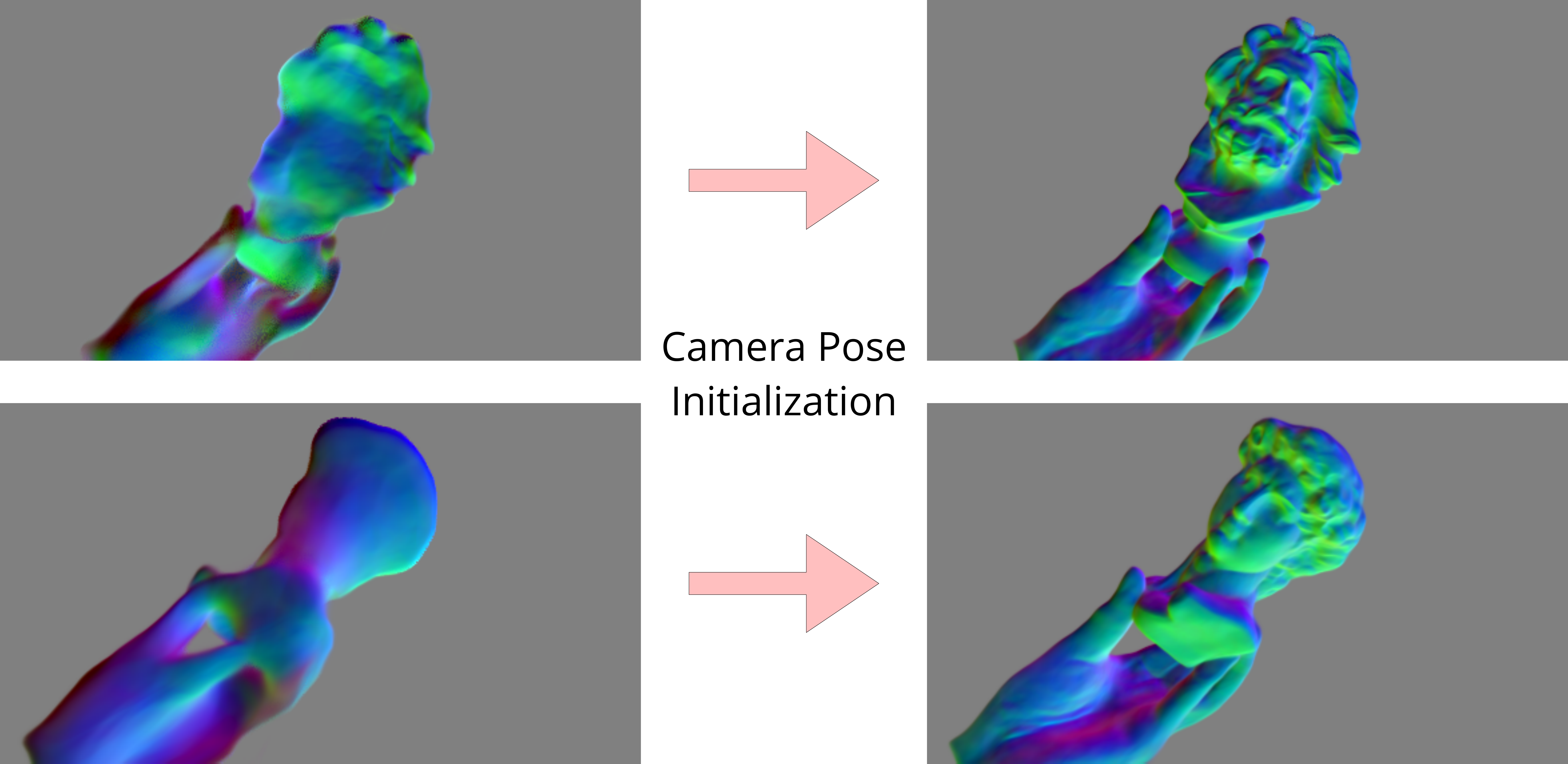}
    \caption{{\textbf{Effectiveness of camera pose initialization.}}
    {\textnormal{We use hand tracking to initialize the camera poses of dense reconstruction.
    Here we compare the reconstructed normal map with and without camera pose initialization.}}
    }
  \label{fig: initialize}
  \end{center}
\end{figure}

We validate our dense reconstruction performance with and without using hand tracking as the camera pose initialization. 
The reconstructed normal map is shown in Figure \ref{fig: initialize}.
Even if our dense reconstruction stage can optimize the camera poses during the optimization phase, it leads to coarse and blurred meshes without camera pose initialization. 
This is because the camera refinement requires a good initialization to achieve correct camera pose refinement. 
With hand tracking, our method is able to generate more detailed object geometry.

\subsection{More results}
\label{sec:appendix more}
\begin{figure*}[hbp]
  \begin{center}
    \includegraphics[width=1.0\linewidth]{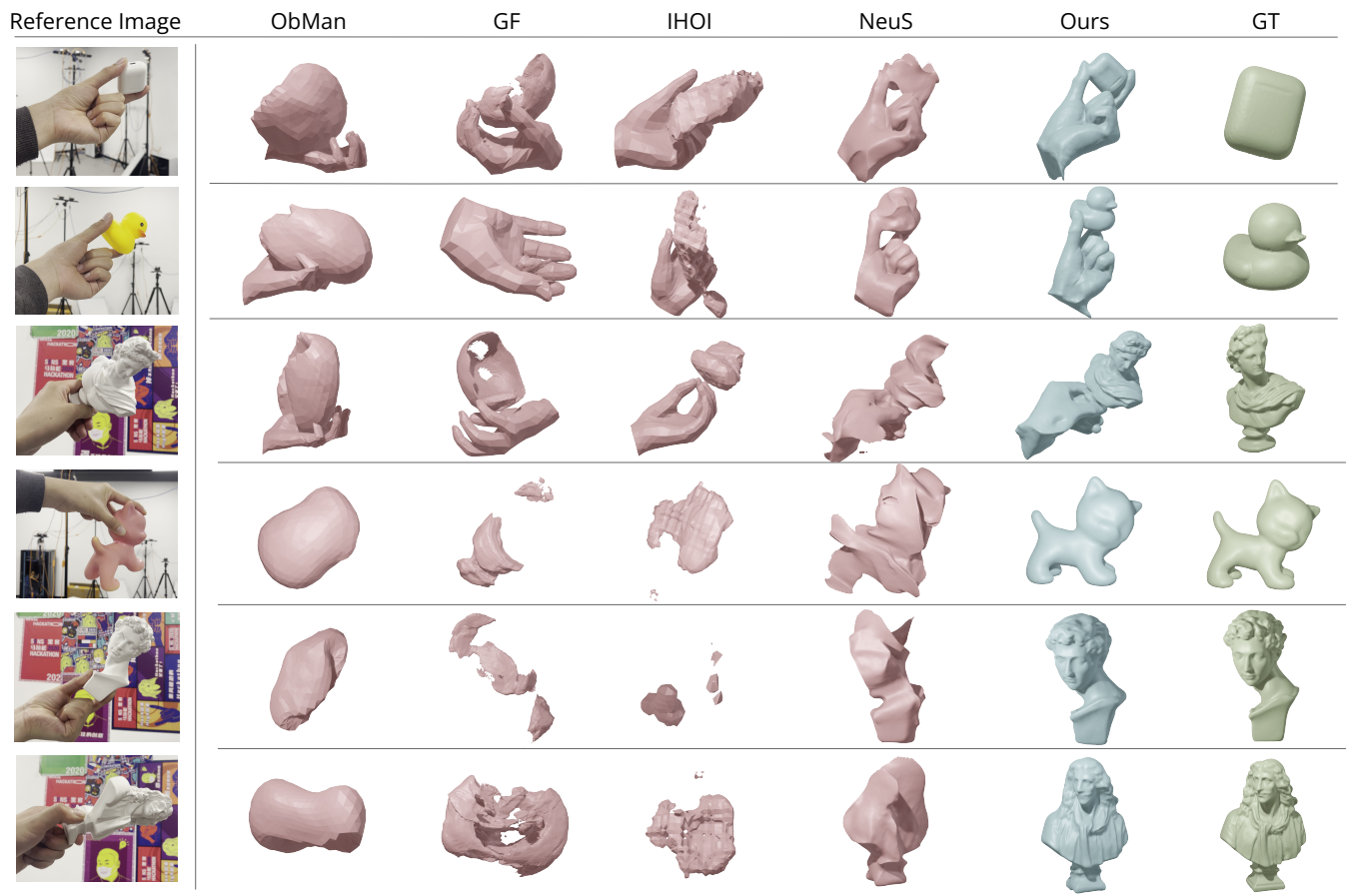}
    \caption{
      {\textbf{A full figure of method comparison.}}
      \textnormal{The top three rows are the joint reconstruction of the hand and object, and the bottom three rows are the separated objects. Learning-based methods ObMan \cite{hasson19_obman}, GF \cite{karunratanakul2020grasping} and IHOI \cite{ye2022hand} cannot generalize to unseen objects; NeuS generates wrong shapes with sharp artifacts. 
      Our method can recover high-quality meshes for both joint hand-object reconstruction and separated object reconstruction.}
    }
  \label{fig:complete}
  \end{center}
\end{figure*}

\begin{figure*}[htp]
  \begin{center}
    \setlength{\abovecaptionskip}{0.1cm}
    \includegraphics[width=1.0\linewidth]{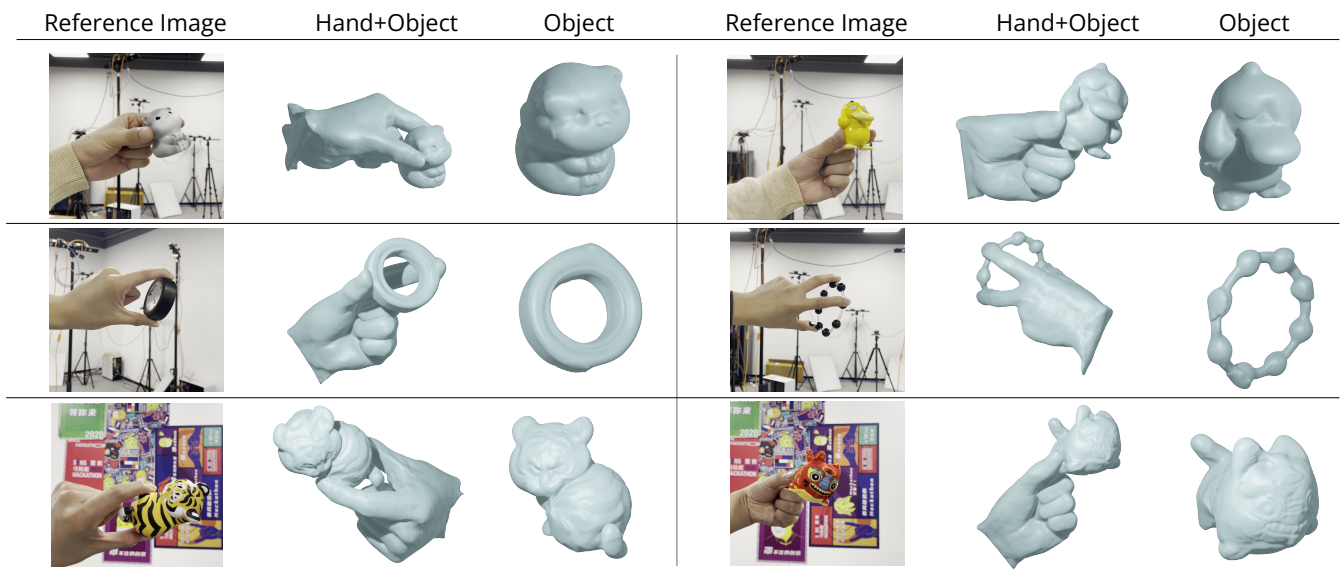}
    \caption{
      {\textbf{More results of our hand-held object reconstruction method.}}
    }
  \label{fig:more}
  \end{center}
\end{figure*}

We give a full figure of method comparison in Figure \ref{fig:complete}, and more reconstructed results in Figure \ref{fig:more}.

\clearpage

\end{document}